\documentclass{article}

\usepackage{arxiv}

\usepackage[utf8]{inputenc} %
\usepackage[T1]{fontenc}    %
\usepackage{hyperref}       %
\usepackage{url}            %
\usepackage{booktabs}       %
\usepackage{amsfonts}       %
\usepackage{nicefrac}       %
\usepackage{microtype}      %
\usepackage{lipsum}         %
\usepackage{graphicx}
\usepackage[square,sort,comma,numbers]{natbib}
\usepackage{doi}

\usepackage{amsmath}
\usepackage{bbm}

\usepackage{comment}
\usepackage[utf8]{inputenc} %
\usepackage[T1]{fontenc}    %
\usepackage{hyperref}       %
\usepackage{url}            %
\usepackage{booktabs}       %
\usepackage{amsfonts}       %
\usepackage{nicefrac}       %
\usepackage{microtype}      %
\usepackage{xcolor}         %
\usepackage{multirow}
\usepackage{array}
\newcolumntype{P}[1]{>{\centering\arraybackslash}p{#1}}
\newcolumntype{M}[1]{>{\centering\arraybackslash}m{#1}}
\usepackage[most]{tcolorbox}
\usepackage{float}
\usepackage{placeins}

\makeatletter
\def\moverlay{\mathpalette\mov@rlay}
\def\mov@rlay#1#2{\leavevmode\vtop{%
   \baselineskip\z@skip \lineskiplimit-\maxdimen
   \ialign{\hfil$\m@th#1##$\hfil\cr#2\crcr}}}
\newcommand{\charfusion}[3][\mathord]{
    #1{\ifx#1\mathop\vphantom{#2}\fi
        \mathpalette\mov@rlay{#2\cr#3}
      }
    \ifx#1\mathop\expandafter\displaylimits\fi}
\makeatother

\newcommand{\bigcupdot}{\charfusion[\mathop]{\bigcup}{\cdot}}

\date{}
\title{Red PANDA: Disambiguating Anomaly Detection by Removing Nuisance Factors}
\usepackage{todonotes} %

\author{%
  Niv Cohen $\quad$  Jonathan Kahana $\quad$ Yedid Hoshen%
  \\
  School of Computer Science and Engineering\\
  The Hebrew University of Jerusalem, Israel\\
}

\hypersetup{
pdftitle={A template for the arxiv style},
pdfsubject={q-bio.NC, q-bio.QM},
pdfauthor={David S.~Hippocampus, Elias D.~Striatum},
pdfkeywords={First keyword, Second Disentanglement, More},
}

\begin{document}

\maketitle

\begin{abstract}
Anomaly detection methods strive to discover patterns that differ from the norm in a semantic way. This goal is ambiguous as a data point differing from the norm by an attribute e.g., age, race or gender, may be considered anomalous by some operators while others may consider this attribute irrelevant. Breaking from previous research, we present a new anomaly detection method that allows operators to exclude an attribute from being considered as relevant for anomaly detection. Our approach then learns representations which do not contain information over the nuisance attributes. Anomaly scoring is performed using a density-based approach. Importantly, our approach does not require specifying the attributes that are relevant for detecting anomalies, which is typically impossible in anomaly detection, but only attributes to ignore. An empirical investigation is presented verifying the effectiveness of our approach.
 
\end{abstract}


\section{Introduction}
Anomaly detection, discovering unusual patterns in data, is a key capability for many machine learning and computer vision applications. In the typical setting, the learner is provided with training data consisting only of normal samples, and is then tasked with classifying new samples as normal or anomalous. It has emerged that the representations used to describe data are key for anomaly detection in images and videos~\cite{reiss2021panda}. Advances in deep representation learning \cite{huh2016makes} have been used to significantly boost anomaly detection performance on standard benchmarks. However, these methods have not specifically addressed biases in data. Anomaly detection methods which suffer from the existence of such biases may produce more overall errors, and incorrectly classify as anomalies some types of samples more than others. A major source for such biases is the presence of additional, nuisance factors. %

One of the most important and unsolved challenges of anomaly detection is resolving the ambiguity between relevant and nuisance attributes. As a motivating example let us consider the application of detecting traffic violations in video. Normal samples consist of videos of usual traffic. When aiming to detect traffic violations, we may encounter two kinds of difficulties: (i) The distribution of anomalous samples is not known at training time e.g. bad driving may come in many forms: speeding, failure to yield, parking in a fire lane, etc. This is the standard problem addressed by most anomaly detection methods \cite{ruff2018deep,reiss2021panda,tack2020csi}. (ii) There may be biases in the normal data. For example, assume that all the taxi drivers in the normal training dataset were females, while all the bus drivers were males. A female driving a bus lawfully, is likely to be considered an anomaly by current methods.

Unlike previous works, we aim to disambiguate between true anomalies (e.g. traffic violations) and unusual variation of nuisance attributes in normal data (e.g. female bus drivers acting lawfully). Detecting normal but unusual variations according to nuisance attributes as anomalies  may be a source of false positive alarms. It may also introduce an undesirable imbalance in the detected anomalies, or even falsely discriminating against certain social groups. The are many settings where some attribute combinations are missing from the training dataset but are considered normal: assembly line training images may be biased in terms of lighting conditions or camera angles - while these may be irrelevant to their anomaly score; photos of people may be biased in terms of ethnicity, for example when collected in specific geographical areas. Moreover, in some cases, normal attribute combinations may be absent just due to the rarity of some attributes (e.g. rare car colors may appear only with specific car models). 

Our technical approach proposes to ignore nuisance attributes by learning representations that are independent from them. Our approach takes as input a training set of normal samples, each labeled with the value of the nuisance attribute that we wish to ignore. Our approach utilizes a domain-supervised disentanglement approach \cite{kahana2022contrastive} to remove the dependency on the provided nuisance attribute, while preserving as much information (uncorrelated to that attribute)  as possible, about the image. 
Specifically, we train an encoder with an additional per-domain contrastive loss term to learn a representation which is independent of the labeled nuisance attribute. %
For example, an encoder guided to be invariant to gender, would be trained to contrast images of females against other images of females, but not against images of males (and vice versa). Additionally, a conditional generator is trained over the representations with a reconstruction term, to ensure the representations are informative.
The combination of the two loss terms yields informative representations which are agnostic to the nuisance attributes. The representations are then combined with standard density estimation methods ($k$ nearest neighbors) for anomaly scoring. %

Our approach differs from previous approaches that propose to use some level of supervision for anomaly detection such as out-of-distribution detection methods or weakly-supervised anomaly detection. Those approaches require knowledge of the true attribute relevant for discriminating anomalies - which is often impossible as anomalies are unexpected. In contrast, we require only knowledge of a subset of the factors that are not indicative for detecting the anomalies we wish to find %
- a far more realistic scenario. In fact, these additional labels are often provided by the datasets, such as self-identified age, gender, or race of employees or customers. In other cases, such labels are easily predicted using pretrained classifiers such as CLIP \cite{radford2021learning}. 

As this task is novel, we present new benchmarks and new metrics for evaluation. Our benchmarks typically %
incorporate normal examples which experience unusual variation in a nuisance attribute. Our evaluation metrics measure both overall anomaly detection accuracy, as well as the false alarm rate due to mistaking normal samples with nuisance variation as anomalies. Our experiments indicate that using our approach for removing the representational dependencies on a nuisance attribute significantly improves both metrics.

\textbf{Contributions:} \textbf{(i)} Introducing the novel setting of \textit{Negative Attribute Guided Anomaly Detection} (NAGAD). \textbf{(ii)}  Presenting new evaluation benchmarks and metrics for the NAGAD setting \textbf{(iii)} Proposing a new approach, \textit{REpresentation
    Disentanglement for Pretrained Anomaly Detection Adaptation} (Red PANDA), using domain-supervised disentanglement to address this setting.
    \textbf{(iv)} Demonstrating the success of our approach through empirical evaluation.

\section{Related Works}

\textbf{Classical anomaly detection methods.} %
These may be grouped into three themes:
(i) \textit{Density-estimation based methods.} %
Estimation of the density of the normal data can be non-parametric methods, such as $k$NN or kernel density estimation. %
Parametric methods, such as Gaussian Mixture Models (GMM) \cite{li2016anomaly} learn a parametric representation of the data to estimate the probability density of the test samples. %
(ii) \textit{Reconstruction based methods} - methods such as PCA learn to reconstruct well normal training samples. Anomalies coming from a different distribution might not reconstruct as well. (iii) \textit{One class classification methods} - Learning a classifier to separate between the train normal samples and the rest of feature space (e.g. SVDD \cite{tax2004support}).  

\textbf{Deep anomaly detection methods.} As only normal samples are available during training, we cannot learn features with standard supervision. Therefore, deep anomaly detection methods either use self-supervision learning to score the anomalies \cite{hendrycks2019using}, or adapt a pretrained representation \cite{hendrycks2019using, reiss2021panda, reiss2021mean, ruff2018deep, perera2019learning} to describe the normal training data.
(i) \textit{Self-supervised methods} - %
these methods learn to solve an auxiliary task on the normal samples, test the performance on new images, and score anomalies accordingly: the network is expected to perform better on the normal samples that come from a similar distribution \cite{hendrycks2019using}. More recent works such as  CSI \cite{tack2020csi} or DROC \cite{goyal2020drocc} use contrastive learning to learn a representation of the normal data. 
(ii) \textit{Adaptation of Pretrained Feature -} Transfer learning of pretrained features was shown to give strong results for out of distribution detection by \cite{hendrycks2019using}. Adaptation of pretrained features for anomaly detection was attempted by Deep-SVDD \cite{ruff2018deep}, which adapted features learnt by an auto-encoder using compactness loss. %
Perera \& Patel suggested to training the compactness loss jointly with ImageNet classification \cite{perera2019learning}. By incorporating early stopping and EWC regularization \cite{kirkpatrick2017overcoming}, PANDA \cite{reiss2021panda} allowed feature adaptation without with mitigated catastrophic forgetting, resulting in better performance. Further improvement in pretrained feature adaptation was later suggested by MeanShifted \cite{reiss2021mean}, using contrastive learning to adapt the pretrained features to the normal training set.

\textbf{Domain-supervised disentanglement.}
Disentanglement is the process of recovering the latent factors that are responsible for the variation between samples in a given dataset. For example, from images of human faces we may recover the age of each person, his hair color, eye color, etc. In domain-supervised disentanglement, one assumes that a single such factor is labelled and aims to learn a representation of the other attributes independent of the labelled factor. This task was approached with variational auto-encoders \cite{jha2018disentangling, bouchacourt2018multi}, and latent optimization \cite{gabbay2019demystifying, gabbay2021scaling}. Contrastive methods have also shown great promise with general disentanglement \cite{zimmermann2021contrastive}. %
This was followed by Kahana \& Hoshen in domain-supervised disentanglement~\cite{kahana2022contrastive} who employed a contrastive loss for each set of similarly-labelled samples individually, learning a code which ideally describes only (and all) attributes which are uncorrelated to the labelled attributes. Domain-supervised disentanglement has been used for a variety of applications. Most notably, for generative models \cite{zhu2018visual}\cite{gabbay2019demystifying}. Self-supervised models have also been discussed in the context of interpretability \cite{hsu2017unsupervised}, abstract reasoning \cite{van2019disentangled}, domain adaptation \cite{peng2019domain}, and fairness \cite{creager2019flexibly}. Some previous works have considered using domain supervision for increasing fairness in anomaly detection\cite{davidson2020framework, zhang2021towards, shekhar2021fairod}. These methods aim at obtaining equal anomaly detection performance across the protected attributes. On the other hand our objective is to ignore the nuisance attributes in order improve the overall performance of the anomaly detection method.

\section{Nuisance Attributes Mislead Anomaly Detectors}
Anomaly detection methods aim to detect samples deviating from the norm. However, operators of anomaly detection methods expect the deviation to be semantically \textit{relevant}. As the anomaly detection setting is typically unsupervised, algorithms are not given guidance as to which mode of deviation is relevant and which is simply nuisance. Detecting anomalies via nuisance attributes is highly undesirable. For example, assume that both pose and car types are the data generating attributes - but pose is nuisance and car type is relevant. Images that different from the norrmal in the pose attribute but not the car model are likely to result in false positive detections. 

Current algorithms rely on different inductive biases to select the relevant attributes and remove the nuisance ones. The most common choice is manual feature selection, where the operator specifies particular features that would be the most relevant. Automated methods for learning features perform a similar function. Contrastive learning methods specify augmentations which remove specific attributes (minor color and location information) from the representation which are considered nuisance. This helps to select attributes more relevant to object-centric tasks. Similarly, representations pretrained on supervised object classifcation (e.g. ImageNet \cite{deng2009imagenet}), which have recently demonstrated very strong results for image anomaly detection, select object-centric attributes at the expense of others. The most extreme level of supervision is the out-of-distribution detection setting where the relevant attribute is labeled for all normal training data. However, this guidance is not available in the typical anomaly detection setting as anomalies are unexpected. e attributes they wish to exclude for anomaly140
detection; either due to legal and moral reasons, 

Our novel setting, \textit{Negative Attribute Guided Anomaly Detection} (NAGAD), allows specification of nuisance attributes which should be ignored by the anomaly detector. Differently from specifying the relevant attributes, which is not possible in anomaly detection, specifying nuisance attributes is often possible. Users may know in advance about the attributes they wish to exclude for anomaly detection; either due to legal and moral reasons, or due to prior domain knowledge.

\begin{figure}[t!]
\centering
\begin{tabular}{P{2cm}c@{\hskip4pt}c@{\hskip4pt}c@{\hskip40pt}c}
& Boots & Sandals & Shoes & Slippers (Anom.) \\ \textbf{Photos} &
\raisebox{-\height/2}{\includegraphics[width=0.15\linewidth]{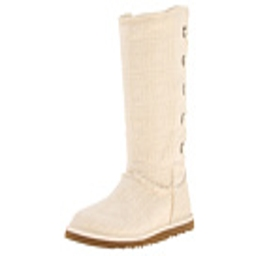}} & 
\raisebox{-\height/2}{\tcbox[colframe=green!30!black, colback=green!30, bottom=0pt, top=0pt, left=0pt, right=0pt]{\includegraphics[width=0.15\linewidth]{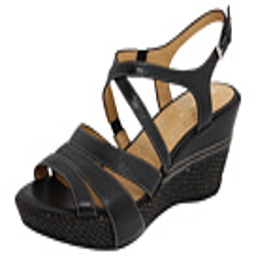}}} &
\raisebox{-\height/2}{\includegraphics[width=0.15\linewidth]{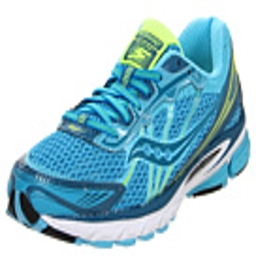}} &
\raisebox{-\height/2}{\tcbox[colframe=red!30!black, colback=red!30, bottom=0pt, top=0pt, left=0pt, right=0pt]{\includegraphics[width=0.15\linewidth]{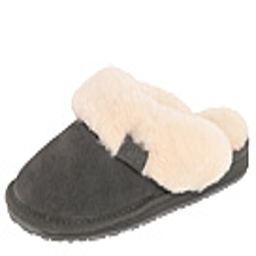}}}
\\
\textbf{Sketches} &
\raisebox{-\height/2}{\tcbox[colframe=green!30!black, colback=green!30, bottom=0pt, top=0pt, left=0pt, right=0pt]{\includegraphics[width=0.15\linewidth]{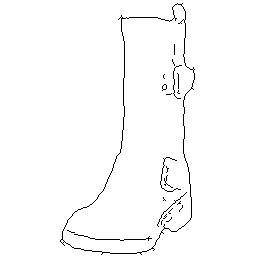}}} & 
\raisebox{-\height/2}{\includegraphics[width=0.15\linewidth]{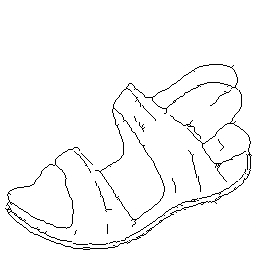}} &
\raisebox{-\height/2}{\includegraphics[width=0.15\linewidth]{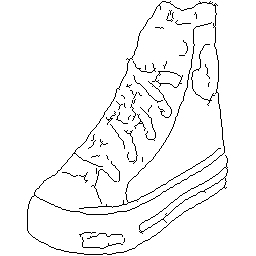}} &
\raisebox{-\height/2}{\tcbox[colframe=red!30!black, colback=red!30, bottom=0pt, top=0pt, left=0pt, right=0pt]{\includegraphics[width=0.15\linewidth]{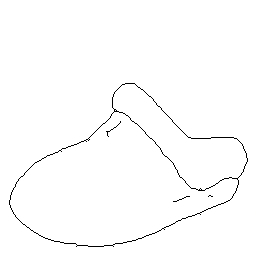}}} \\ 

\end{tabular}
\caption{Samples from the Edges2Shoes datasets. Pseudo-anomalies are marked in green while true anomalies are marked in red. Both pseudo-anomalies and true anomalies appear only in the test set.}
\label{fig:samples}
\end{figure}

A natural way for specifying nuisance attributes is to provide labels for their different classes. For example, wishing to detect anomalies according to their type of shoe but not according to the image type, we may provide for each image a label for the image type (such as sketches vs. photos labels, see Fig.\ref{fig:samples}). Currently available anomaly detection approaches cannot directly benefit from such information and thus mitigate nuisance attributes only implicitly (using the mechanisms explained above). In Sec.~\ref{sec:method} we describe a specific technical approach for using the guidance for anomaly detection. However, we stress that our main contribution is this anomaly detection setting which we expect to significantly reduce false alarms in many cases.

\section{Red PANDA: Disentanglement Approach for Removing a Nuisance Factor}
\label{sec:method}
\subsection{Obtaining Labels for the Nuisance Attribute}
Our approach, \textit{REpresentation
Disentanglement for Pretrained Anomaly Detection Adaptation} (Red PANDA), aims to achieve a representation invariant to a nuisance attribute of our dataset, leading to better detection of anomalies expressed in relevant attributes. To do so, we provide labels for the nuisance attribute. For example, when we wish to detect anomalies in drivers behaviour, we may consider the gender of the driver as a nuisance attribute. Therefore, we wish not to consider the gender attribute in our algorithm during anomaly detection and provide labels for it. 

We have a few options to achieve these labels. In some cases they may already exist in the dataset. A very natural such case is when we have data from a few static cameras, and wish to ignore the camera identity. In many other cases, a pretrained classifier, already trained for these specific attributes may provide such labels. Recently, pretrained models for text-based zero-shot classification such as CLIP \cite{radford2021learning} have shown promising results. They allow to supply of-the-shelf automatic labels for a very large set of attributes. We conducted a small experiment over the \textit{Edges2Shoes} \cite{edges2shoes} dataset, automatically labelling it with CLIP, and achieved $99.97\%$ accuracy in labelling whether an image is a photo or a sketch.
Taken together, although in some cases collecting labels for nuisance attributes may be laborious, in many cases they can be achieved at virtually no cost.

\subsection{Preliminaries}
In our setting, the training set consists of normal samples only denoted as %
${\cal{D}}_{train}$. For each normal image $x_i \in {\cal{D}}_{train}$ we are also provided with its label $n_i$ describing the nuisance attribute we wish to ignore. Our evaluation set ${\cal{D}}_{test}$ consists of both normal and anomalous samples. We denote the normal/anomaly label for a test image $x_i$ as $y_i$. %
For each such dataset, each sample is described by multiple attribute labels $(n_i, a_i, b_i, c_i, ...) \in N \times A \times B \times C \times ...$, where $N$ describe our nuisance attribute, and $A, B, C, ...$ describe different relevant attributes (consider for example the identity of the object, the lightning condition, and camera angles as different attributes). We assume that the anomaly label is always a function of (potentially) all the relevant attributes $y_i = f_a(a_i, b_i, c_i, ...)$. Namely, we assume the nuisance attribute $n_i$ never affects the anomaly label $y_i$. We emphasize that in our described setting, none of the relevant attribute labels nor the anomaly labels are given during training.

We aim to learn an encoder function $f$ mapping samples $x_i$ to a code describing their relevant attributes $f(x_i) \in R^{d}$. We also wish our codes to be aligned. This is, we wish our encoder to represent the relevant attributes in a way which is not affected by the nuisance attributes: 

\begin{equation}
(a_i, b_i, c_i, ...) = (a_j, b_j, c_j, ...) \leftrightarrow f(x_i) \approx f(x_j)
\label{eq:alignment}
\end{equation}
We also need our code to be informative - to represent sufficient information regarding our relevant attributes:

\begin{equation}
I\big((a_i, b_i, c_i, ...); x_i\big) \approx I\big((a_i, b_i, c_i, ...); f(x_i)\big)
\label{eq:informativeness}
\end{equation}

Given such a representation we may later score anomalies independently from any biases caused by the nuisance attribute we wish to ignore.

\subsection{Contrastive Disentanglement}

In this section, we describe the technical approach we employ for ensuring that $f$ does not contain information on the nuisance attribute, while retaining as much information about the relevant attributes \cite{kahana2022contrastive}.

\textbf{Pretrained encoder.} We initialize the encoder function $f$ with an ImageNet pretrained network. ImageNet-pretrained representations were previously shown to be very effective for image anomaly detection\cite{reiss2021panda}. Off-the-self pretrained representation, however, also encodes much information on the nuisance attributes. Therefore by themselves they \textit{do not} satisfy our disentanglement objective.

\textbf{Contrastive loss.} Our objective is that images that have similar relevant attributes but different nuisance attributes would have similar representations. Although we are not provided with supervised matching pairs, we use the proxy objective requiring the distribution of representations of images having different nuisance attributes to be the same \cite{kahana2022contrastive}. To match the distributions we first split our training data ${\cal{D}}_{train}$ to disjoint subsets $S_{n_i}$ according to the nuisance attribute values:
\begin{equation}
{\cal{D}}_{train} = \bigcupdot_{n_i\in N} S_{n_i}
\label{eq:sets}
\end{equation}

We then employ a contrastive loss, on each of the sets $S_{n_i}$ independently:

\begin{equation}
\mathcal{L}_{con} = \log \sum_{{\cal{D}}_{train}, N }^{} \mathbbm{1}_{(n_i = n_j)}e^{sim\big((f(x_i), f(x_j)\big)}
\label{eq:sets}
\end{equation}

This objective encourages the encoder to map the image distribution uniformly to the unit sphere (see Wang and Isola \cite{wang2020understanding}), and therefore is likely to match the marginal distribution $F$ of latent codes across the nuisance attribute. Specifically, we would like the distribution of encoded features $F$ to be independent of the nuisance attribute $n_i$: $p(F|n_i) = p(F|n_j)$. We note that matching of marginal distributions is necessary, but not a sufficient condition for alignment (Eq.~\ref{eq:alignment}). Yet, this often appears to happen in practice. 

Another problem that may arise is insufficient informativeness: the contrastive objective does not prevent ignoring some of the relevant attributes \cite{chen2021intriguing}. To support the informativeness we add an augmentation loss, encouraging different augmentations of the same image to me mapped to similar codes: $\mathcal{L}_{aug} = -sim\Big(f\big(A_1(x_i),A_2(x_i)\big)\Big)$. To further encourage informativeness, we also employ a reconstruction loss.

\textbf{Reconstruction loss.} To require the representation to contain as much information about the relevant attributes as possible, we use a reconstruction constraint. Specifically, we require that given the combination of the representation $f_i$ (which ideally ignores the nuisance attribute) and the value of the nuisance attribute $n_i$, it should be possible to perfectly recover the sample $x_i$. This is enforced using a generator function $G$ which performs this as a regression task. The generator is trained end-to-end together with the encoder. The reconstruction is measured using a perceptual loss. %

\begin{equation}
\mathcal{L}_{rec} = \sum_{{\cal{D}}_{train}, N }^{}\ell_{perceptual}\Big( x_i, G \big(f(x_i), n_i\big)\Big)
\label{eq:sets}
\end{equation}

\subsection{Anomaly Scoring}
In the previous steps we learned an encoder $f$ that maps each image into a compact representation of its relevant attributes. In this section, we estimate the probability distribution of the normal data in the representation space for anomaly detection. Similarly to other anomaly detection methods, we hypothesize the anomalous samples will be mapped to low-density regions, while normal data will be mapped to high-density regions. This assumption is violated in the case where the representation contains both relevant and nuisance attributes; as unusual combinations of relevant and nuisance attributes will be rare and therefore classified as anomalous. However, if the representation contains only relevant attributes, low-density regions would indeed correspond to samples with rare relevant attributes - which are indeed likely to be anomalous. 

To numerically estimate the density of the normal data around each test sample, we use the $k$ nearest neighbours algorithm ($k$NN). We begin with extracting the representation for each normal samples: $ f_i = f(x_i), \quad \forall x_i	\in  {\cal{D}}_{train}$. Next, for each test sample we infer its latent code $f_t = f(x_t)$. Finally, we score it by the $k$NN distance to the normal data:

\begin{equation}
 S(x_t) = \sum_{f_i \in N_k(f_t)}^{} sim(f_i, f_t) 
\end{equation}
where $N_k(f_t)$ denotes the $k$ most similar relevant attribute feature vectors in the normal data. We note that although we trained our encoder $f$ with a contrastive loss, encouraging uniform distribution in the sphere, the high dimension of the latent space allows us to distinguish between high and low density areas of the distribution of normal data.

\textbf{Runtime.} Although $k$NN has runtime complexity linear in the number of training data, it can be sped up using K means or core-set techniques (as done in SPADE \cite{cohen2020sub} or PatchCore \cite{roth2021towards}). In practice, the wall-clock runtime of the retrieval stage of our approach is minimal, even without such speedups (>3500 images per second for the SmallNorb dataset).   

 \begin{figure}[b!]
\centering
\begin{tabular}{P{1.2cm}c@{\hskip4pt}c@{\hskip4pt}cc@{\hskip20pt}c}
& Type-1 & Type-2 & Type-3 & Type-4 & Type-5 (Anom.) \\ \textbf{Pose 1} &
\raisebox{-\height/2}{\includegraphics[width=0.1\linewidth]{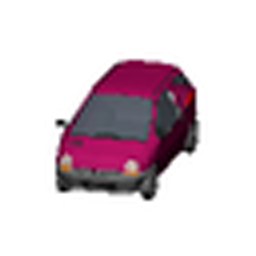}} & 
\raisebox{-\height/2}{\includegraphics[width=0.1\linewidth]{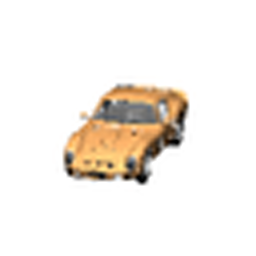}} &
\raisebox{-\height/2}{\includegraphics[width=0.1\linewidth]{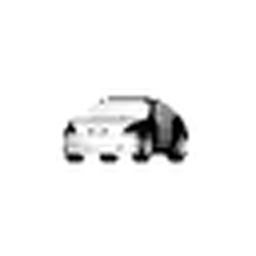}} &
\raisebox{-\height/2}{\tcbox[colframe=green!30!black, colback=green!30, bottom=0pt, top=0pt, left=0pt, right=0pt]{\includegraphics[width=0.1\linewidth]{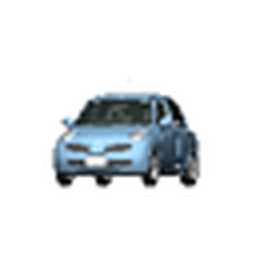}}} &
\raisebox{-\height/2}{\tcbox[colframe=red!30!black, colback=red!30, bottom=0pt, top=0pt, left=0pt, right=0pt]{\includegraphics[width=0.1\linewidth]{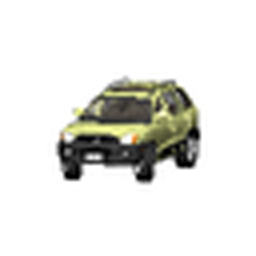}}}
\\ 
\textbf{Pose 4} &
\raisebox{-\height/2}{\includegraphics[width=0.1\linewidth]{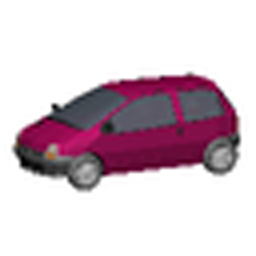}} & 
\raisebox{-\height/2}{\includegraphics[width=0.1\linewidth]{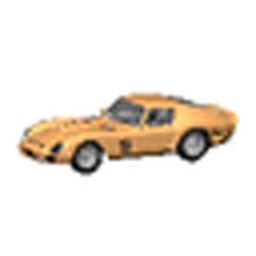}} &
\raisebox{-\height/2}{\tcbox[colframe=green!30!black, colback=green!30, bottom=0pt, top=0pt, left=0pt, right=0pt]{\includegraphics[width=0.1\linewidth]{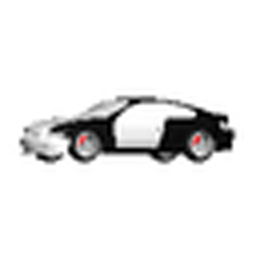}}} &
\raisebox{-\height/2}{\includegraphics[width=0.1\linewidth]{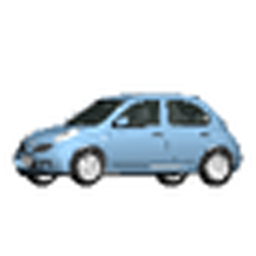}} &
\raisebox{-\height/2}{\tcbox[colframe=red!30!black, colback=red!30, bottom=0pt, top=0pt, left=0pt, right=0pt]{\includegraphics[width=0.1\linewidth]{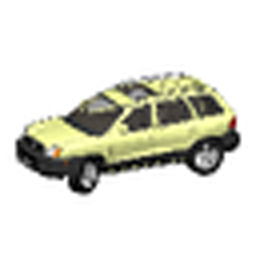}}}
\\ 
\textbf{Pose 9} &
\raisebox{-\height/2}{\includegraphics[width=0.1\linewidth]{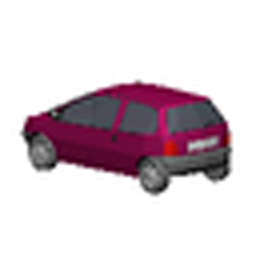}} & 
\raisebox{-\height/2}{\tcbox[colframe=green!30!black, colback=green!30, bottom=0pt, top=0pt, left=0pt, right=0pt]{\includegraphics[width=0.1\linewidth]{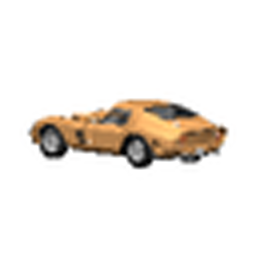}}} &
\raisebox{-\height/2}{\includegraphics[width=0.1\linewidth]{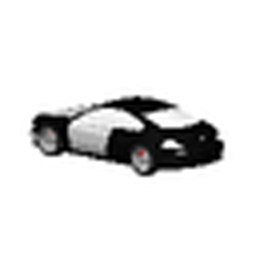}} &
\raisebox{-\height/2}{\includegraphics[width=0.1\linewidth]{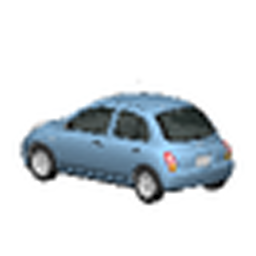}} &
\raisebox{-\height/2}{\tcbox[colframe=red!30!black, colback=red!30, bottom=0pt, top=0pt, left=0pt, right=0pt]{\includegraphics[width=0.1\linewidth]{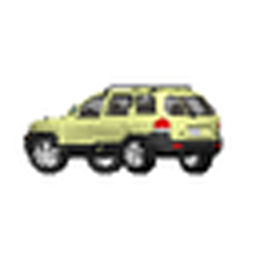}}}
\\ 
\textbf{Pose 17} &
\raisebox{-\height/2}{\tcbox[colframe=green!30!black, colback=green!30, bottom=0pt, top=0pt, left=0pt, right=0pt]{\includegraphics[width=0.1\linewidth]{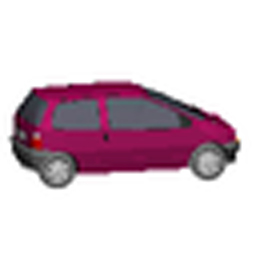}}} & 
\raisebox{-\height/2}{\includegraphics[width=0.1\linewidth]{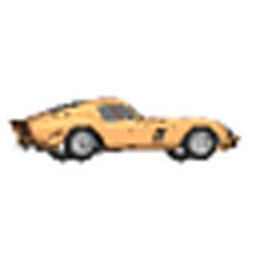}} &
\raisebox{-\height/2}{\includegraphics[width=0.1\linewidth]{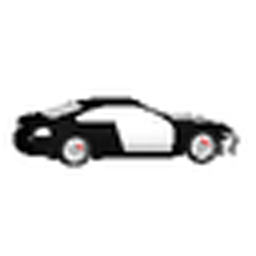}} &
\raisebox{-\height/2}{\includegraphics[width=0.1\linewidth]{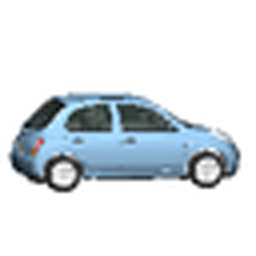}} &
\raisebox{-\height/2}{\tcbox[colframe=red!30!black, colback=red!30, bottom=0pt, top=0pt, left=0pt, right=0pt]{\includegraphics[width=0.1\linewidth]{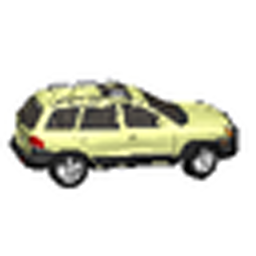}}}

\end{tabular}
\caption{Samples from the Cars3D datasets. Pseudo-anomalies are marked in green while true anomalies are marked in red. Both pseudo-anomalies and true anomalies appear only in the test set.}
\label{fig:samples_cars}
\end{figure}

\section{Experiements}
\label{sec:experiements}
\subsection{Setting}

\textbf{Benchmark construction.} As our anomaly detection setting is novel, new benchmarks need to be designed for its evaluation. The following protocol is proposed for creating the benchmarks. First, we select an existing dataset containing multiple labelled attributes. We designate one of its attributes as nuisance, e.g. the object pose, and other attributes as relevant, e.g. the identity of the object. Only the relevant attributes are used to designate an object as anomalous whereas the nuisance attribute does not. We then remove some combinations of nuisance and relevant attributes from the training set, creating bias in the data. For example, we may remove all left-facing cars for one car model, and right-facing cars for another car model. As these combinations of attributes are not present in the normal train set, we refer to them as \textit{pseudo-anomalies}. We refer to any sample that shares all the attributes (including nuisance attributes) with a normal training sample as a \textit{familiar sample}. In this setting, we aim both to both detect true anomalies (anomalies according to the relevant attributes), and treat pseudo-anomalies as normal as the familiar-samples, as they differ from the normal data only in nuisance attributes,

\textbf{Metrics.} In our setting, we wish not only to measure our overall anomaly detection performance, but also to evaluate the false alarm rate due to pseudo-anomalies. We therefore report our results in terms of three different scores. Each such score uses two subsets of the test set, and measures by ROC-AUC how well does our anomaly detection score distinguish between them: (i) Standard anomaly detection (AD)-Score, which measures how accurately anomalies are detected with respect to the normal test data (both seen combinations and pseudo anomalies). (ii) Pseudo anomalies (PA)-Score: measures how much pseudo-anomalies are scored as more anomalous than familiar-samples  (iii) Relative abnormality (RA)-score: measures how accurately true anomalies are detected compared to pseudo-anomalies. Taken together, these metrics measure the accuracy of an anomaly detector while not being biased by unseen combinations of nuisance attributes.

\subsection{Results}

We report the results on three multi-attribute datasets based on Cars3D, SmallNORB and Edges2Shoes. We chose these specific datasets as they are the common datasets in the field of domain-supervised disentanglement \cite{gabbay2019demystifying, kahana2022contrastive}. We find these datasets to be also non-trivial for state-of-the-art anomaly detection algorithms. %

\textbf{Compared Methods.}

\textit{DN2 \cite{reiss2021panda}.} A simple but effective approach fully reliant on pertaining. It uses an ImageNet-pretrained network to extract representations for each image. Each test image is scored using $k$NN density estimation similarly to our approach.

\textit{MeanShifted \cite{reiss2021mean}.} A recent method that achieves state-of-the-art performance on standard anomaly benchmarks. It uses a modified contrastive learning loss to adapt its feature to the normal train set. This method uses the same pretrained network as our method to initialize the features. It then uses a $k$NN for anomaly scoring.

\textit{CSI \cite{tack2020csi}.} A strong self-supervised anomaly detection method that does not rely on pretraining. It uses two types of augmentations, fine changes simulating positive contrastive loss samples, and domain shifts as negative samples. Anomaly scoring is performed using an ensemble of similarity scores based on the learnt features. 

\textit{SimCLR \cite{chen2020simple}.} An ablation of our approach that trains a single contrastive loss rather than a different contrastive loss for each domain. We score the anomalies similarly to our approach. 

\textbf{Evaluation.} For each dataset, we label each sample as either \textit{normal}, \textit{true anomaly}, or \textit{pseudo-anomaly} as detailed below. We include true anomalies and pseudo-anomalies only in the test set, and split the normal samples between the training set and the test set ($85\% / 15\%$ train/test split).

\textbf{Datasets.}

\textit{Cars3D} \cite{carsdataset}. A synthetic image dataset, with each image formed from two attributes: car model and pose. Car models are varied across different colors, shapes and functionalities. Each car model is observed from multiple camera angles (pose). We define true anomalies as $5$ (randomly selected) car models. To simulate pseudo anomalies, we randomized for each camera angle another single car model and labeled it as \textit{pseudo-anomaly}. An illustration of the dataset can be seen in Fig. \ref{fig:samples_cars}.

We can see in Tab.\ref{tab::cars3d} that the disentanglement approach significantly outperforms methods that do not use any guidance to remove the nuisance attribute. The method detects true anomalies, without assigning high anomaly scores to the pseudo-anomalies, significantly better than all other methods compared. The RA-Score shows that our detector scores true anomalies significantly higher than pseudo-anomalies.

\begin{table}[!ht]
    \centering
    \caption{Empirical Evaluation on the Cars3D Dataset (ROC-AUC) }
    \label{tab::cars3d}
    \begin{tabular}{P{2cm}P{3.5cm}P{2cm}P{2cm}P{2cm}}
    \toprule
        \textbf{Dataset} & \textbf{Method} & \textbf{AD-Score ($\Uparrow$)} & \textbf{PA-Score ($\Downarrow$)} & \textbf{RA-Score ($\Uparrow$)} \\
        \midrule
        \multirow{6}{*}{\textbf{Cars3D}}
        & SimCLR & 0.780 & 0.519 & 0.741 \\
        & CSI & 0.606 & 0.579 & 0.538 \\ 
        & DN2 & 0.946 & 0.564 & 0.916 \\ 
        & MeanShifted & 0.943 & 0.595 & 0.917 \\ 
        \cmidrule(lr){2-5}
        & Ours & \textbf{0.985} & \textbf{0.506} & \textbf{0.980} \\ 
        \bottomrule
    \end{tabular}
\end{table}

\textit{SmallNorb \cite{smallNORB}.} Each image is synthetically constructed from several attributes: object type, camera azimuth, camera elevation and lighting. The object types come from different categories such as animals, people, planes, trucks and cars. To simulate our anomalies we randomized a single object class (e.g. deer) from each category type. We define the camera azimuth angles as our nuisance attribute. For each azimuth angle value we randomize a single object class, and assign samples of that type and camera angle as pseudo-anomalies.

We can see in Tab.\ref{tab::norb} that our approach outperforms the baselines on all metrics. All the methods utilizing pretrained features detect true anomalies fairly well. This could be expected, as during pretraining the network learns a good representation of objects. Our disentanglement approach significantly reduces the tendency to score pseudo-anomalies as anomalies. %
CSI treats pseudo-anomalies similarly to normal samples, but this is most likely because its representation for this dataset is not informative, and does not distinguish well between unseen data (be it true anomalies or pseudo-anomalies) and the rest of the test data.

\begin{table}[!ht]
    \centering
    \caption{Empirical Evaluation on the SmallNorb Dataset (ROC-AUC)}
    \label{tab::norb}
    \begin{tabular}{P{2cm}P{3.5cm}P{2cm}P{2cm}P{2cm}}
    \toprule
        \textbf{Dataset} & \textbf{Method} & \textbf{AD-Score ($\Uparrow$)} & \textbf{PA-Score ($\Downarrow$)} & \textbf{RA-Score ($\Uparrow$)} \\
        \midrule
        \multirow{6}{*}{\textbf{SmallNorb}}
        & SimCLR & 0.805 & 0.728 & 0.638 \\ 
        & CSI & 0.618 & \textbf{0.556} & 0.575 \\ 
        & DN2 & 0.908 & 0.819 & 0.768 \\ 
        & MeanShifted & 0.948 & 0.870 & 0.811 \\ 
        \cmidrule(lr){2-5}
        & Ours & \textbf{0.953} & 0.581 & \textbf{0.943} \\
        \bottomrule
    \end{tabular}
\end{table}

\textit{Edges2Shoes \cite{edges2shoes}.} This dataset contains photos of shoes and edge maps images of the same photos. They are labelled in terms of image type (sketch vs. photo), shoe type, and other attributes (the labels come from the original UT-Zappos50K dataset \cite{zappos50k}). We assign all images with shoe type "slippers" as a true anomaly. We assign all photos of type "sandal", and all sketches of type "boot" as pseudo-anomalies. An illustration of the dataset can be seen in Fig. \ref{fig:samples}.

This dataset is challenging as the photo and sketch domains are quite far, making the nuisance attribute dominant. E.g., by observing only sketches of boots, real photos of boots could be easily considered as anomalies without further guidance. %
Our approach outperforms methods that do not remove nuisance attributes from the representation. We observe (by the PA-score) that although the pseudo-anomalies are indeed scored higher than normal images by our approach, their scores are still higher than the true anomalies (demonstrated by the RA-score). Our approach significantly outperforms the baselines, showcasing the importance of specifying and removing nuisance attributes.

\begin{table}[!ht]
    \centering
    \label{tab::e2s}
    \caption{Empirical Evaluation on the Edges2Shoes Dataset (ROC-AUC) }
    \begin{tabular}{P{2cm}P{3.5cm}P{2cm}P{2cm}P{2cm}}
    \toprule
        \textbf{Dataset} & \textbf{Method} & \textbf{AD-Score ($\Uparrow$)} & \textbf{PA-Score ($\Downarrow$)} & \textbf{RA-Score ($\Uparrow$)} \\
        \midrule
        \multirow{6}{*}{\textbf{Edges2Shoes}}
        & SimCLR & 0.567 & 0.642 & 0.510 \\ 
        & CSI & 0.574 & 0.873 & 0.412 \\ 
        & DN2 & 0.500 & \textbf{0.631} & 0.455 \\ 
        & MeanShifted & 0.486 & 0.790 & 0.386 \\ 
        \cmidrule(lr){2-5}
        & Ours & \textbf{0.781} & 0.711 & \textbf{0.719} \\
        \bottomrule
    \end{tabular}
\end{table}

\section{Discussion}

\textbf{Multi-attribute dataset.} Many datasets (e.g. SmalNorb) have more than two attributes. In some cases, we may wish to remove multiple nuisance attributes. Methods such as \cite{gabbay2019demystifying} very naturally extends to the case of disentangling many factors of the same dataset. We believe that our approach can also be extended to this setting.

\textbf{Supervised vs. self-supervised pretraining.} Many top performing approaches (including ours) rely on externally-pretrained weights for initializing their neural networks. Pretrained weights implicitly provide useful guidance regarding the relevant attributes we should focus on, and the ones we may wish to ignore (e.g. low-level image information).  Different pretrained networks provide different relevant/nuisance attribute splits. We found that pretrained weights obtained from supervised classification on external datasets such as ImageNet, tend to emphasize the main object featured in the center of the image, and are more invariant to other attributes. Representations learned by self-supervised pretraining on external datasets are affected both by the external dataset but also by the augmentation used for its contrastive learning. Therefore they have difference inductive biases.

\textbf{Augmentations.} Different methods may require augmented images to be similar or dissimilar to the original image \cite{chen2020simple, tack2020csi}. This choice tends to have a strong effect on the results. E.g., a network trained to be rotation invariant, may fail when the relevant attribute is image orientation angle. %
Our approach only uses simple augmentations such as Gaussian blurring, saturation and crops. We expect these augmentations not to restrict the anomalies detectable in the vast majority of cases. In general, augmentation should be carefully inspected when deploying anomaly detection methods in practice.

\textbf{Removing nuisance attributes with generative models.} Recently, generative models e.g. StyleGAN \cite{karras2019style} have been able to learn very powerful representations for several data types, particularly images of faces. Their representations experience a certain level of disentanglement \cite{wu2021stylespace}. When available, such models can be utilized for removing nuisance attributes in a similar approach to ours. %

\section{Limitations}
\label{sec:limitations}
\textbf{Domain supervised disentanglement in the wild.} Currently, state-of-the-art domain-supervised disentanglement methods achieve impressive results on synthetic or curated datasets. Such methods do not perform as well for in-the-wild datasets. As our approach heavily relies on disentanglement, it is prone to similar limitations. As the field of disentanglement advances, the advancements can be directly translated to improved anomaly detection in our approach.

\textbf{Highly biased datasets.} Similarly to other disentanglement approaches, we require the distributions of relevant attributes across nuisance domains to be somewhat similar. We have shown that our method can work when the supports across domains are not overlapping. Still, we expect that when the support is highly non-overlapping the results will significantly deteriorate. Developing methods able to disentangle domains with highly non-overlapping support is an exciting future direction.
\section{Conclusion}
We proposed a new anomaly detection setting where information is provided on a set of attributes that are known to be irrelevant from distinguishing normal from anomalous data. Using a disentanglement-based approach, we showed how this additional supervision can be leveraged for better anomaly detection in biased datasets. As knowing a subset of the attributes that are irrelevant is much easier than knowing in advance the entire list of relevant attributes, we expect our new setting to be useful in practical applications.

\section{Acknowledgements}
This work was partly supported by the Malvina and Solomon Pollack scholarship and an Israeli Council for Higher Education grant in the data sciences.

\bibliographystyle{unsrt}
\bibliography{mybib}
\clearpage

\pagenumbering{arabic}%
\renewcommand*{\thepage}{A\arabic{page}}
\section{Appendix}

\subsection{Implementation Details}

\subsubsection{Disentanglement module}

We use most of the parameters as in the DCoDR paper\cite{kahana2022contrastive} for our disentanglement module. All images were used in a $64\times\ 64$ resolution. 
For the contrastive temperature, we use $\tau = 0.1$ for all the datasets.
We scale down the loss $\mathcal{L}_{rec}$ by a factor of $0.3$.

\textbf{Architecture.} A ResNet50 encoder pretrained on image classifications. In accordance with previous works, we add 3 fully-connected layer to the encoder for the SmallNorb dataset \cite{gabbay2019demystifying,kahana2022contrastive}.
For the perceptual loss of the generator we used a VGG network pretrained on ImageNet.

\textbf{Optimization.}
We use $200$ training epochs. In each batch we used $32$ images from $4$ different nuisance classes (a batch size of $128$, in total). We used a learning rates of $1\cdot10^{-4}$ and $3\cdot10^{-4}$ for the encoder and generator (respectively). 

\textbf{Augmentation.} We used Gaussian blurring (kernel\_size $=5$, $\sigma=1$), high contrast (contrast $=(1.8,3.0)$), and high saturation (saturation $=(1.8,3.0)$) for our augmentation. For Edges2Shoes we used only Gaussian Blurring. For the SimCLR \cite{chen2020simple} contrastive learning (both in our approach and the baseline), we follow DCoDR by only augmenting the original image once, and comparing the augmented and the original views encodings. This in contrast to SimCLR which compares two augmented views instead.

\subsubsection{Scoring module}
 We use \textit{faiss}\cite{johnson2019billion} $k$NN implementation, using $k = 1$.
As our similarity measure we use Cosine distance, similarly to the distance used during our contrastive training.

\subsubsection{Datasets}

To simulate anomalies in the dataset, we first designate true anomalies as described in Sec.\ref{sec:experiements}. We then chose combination of normal classes and the nuisance attribute to designate pseudo anomalies. We used the following random combinations for pseudo anomalies:

\textit{Cars3D:}

\begin{table}[!ht]
    \centering
    \label{tab::norb_psuedo}
    \caption{Cars3D Psuedo Anomalies Selection}
    \begin{tabular}{P{2cm}P{2cm} | P{2cm}P{2cm}}
    \toprule
        \textbf{Azimuth} & \textbf{Object Type} & \textbf{Azimuth} & \textbf{Object Type} \\
        \midrule
        \textbf{0} & 173 & \textbf{12} & 48 \\
        \textbf{1} & 16 & \textbf{13} & 66 \\
        \textbf{2} & 75 & \textbf{14} & 32 \\
        \textbf{3} & 23 & \textbf{15} & 153 \\
        \textbf{4} & 44 & \textbf{16} & 128 \\
        \textbf{5} & 78 & \textbf{17} & 120 \\
        \textbf{6} & 108 & \textbf{18} & 38 \\
        \textbf{7} & 7 & \textbf{19} & 172 \\
        \textbf{8} & 167 & \textbf{20} & 106 \\
        \textbf{9} & 182 & \textbf{21} & 4 \\
        \textbf{10} & 99 & \textbf{22} & 175 \\
        \textbf{11} & 78 & \textbf{23} & 111 \\
        \bottomrule
    \end{tabular}
\end{table}

 \clearpage

\textit{SmallNorb:}

\begin{table}[!ht]
    \centering
    \label{tab::norb_psuedo}
    \caption{Smallnorb Psuedo Anomalies Selection}
    \begin{tabular}{P{2cm}P{2cm} | P{2cm}P{2cm}}
    \toprule
        \textbf{Azimuth} & \textbf{Object Type} & \textbf{Azimuth} & \textbf{Object Type} \\
        \midrule
        \textbf{0} & 44 & \textbf{9} & 38 \\ 
        \textbf{1} & 17 & \textbf{10} & 35 \\
        \textbf{2} & 9 & \textbf{11} & 12 \\
        \textbf{3} & 25 & \textbf{12} & 24 \\
        \textbf{4} & 48 & \textbf{13} & 35 \\
        \textbf{5} & 20 & \textbf{14} & 29 \\
        \textbf{6} & 12 & \textbf{15} & 23 \\
        \textbf{7} & 44 & \textbf{16} & 41 \\
        \textbf{8} & 8 & \textbf{17} & 43 \\
        \bottomrule
    \end{tabular}
\end{table}

\textit{Edges2Shoes:}

\begin{table}[!ht]
    \centering
    \label{tab::norb_psuedo}
    \caption{Edges2Shoes Psuedo Anomalies Selection}
    \begin{tabular}{P{2cm}P{2cm}}
    \toprule
        \textbf{Image Type} & \textbf{Shoe Type} \\
        \midrule
        Photo &  Sandals \\
        Sketch &  Boots \\
        \bottomrule
    \end{tabular}
\end{table}
\FloatBarrier

Finally we take all of the psuedo anomalies to the test set.

\subsection{Compute Resources}

The entire project used in total $3000$ hours of NVIDIA RTX A5000 GPU (including development, testing and comparisons). %
All resources were supplied by a local internal cluster.

\subsection{Typical Statistical Error in Experimental Results}

As our experiments are relatively long and results are fairly consistent among different runs we do not provide an error bar for each single run. As a typical case, we ran $3$ repetitions of our approach for the SmallNorb experiments. The consistency of the results is presented in Tab.\ref{tab::stat_norb}.

\begin{table}[!ht]
    \centering
    \caption{Consistency of results among repetitions for the SmallNorb dataset (ROC-AUC)}
    \label{tab::stat_norb}
    \begin{tabular}{P{2cm}P{2cm}P{2cm}P{2cm}}
    \toprule
        \textbf{Dataset}  & \textbf{AD-Score ($\Uparrow$)} & \textbf{PA-Score ($\Downarrow$)} & \textbf{RA-Score ($\Uparrow$)} \\
        \midrule
        \multirow{1}{*}{\textbf{SmallNorb}}
         & 0.952 $\pm$ 0.008 & 0.553 $\pm$ 0.015 & 0.947  $\pm$ 0.008 \\
        \bottomrule
    \end{tabular}
\end{table}

\subsection{License}

Our technical approach is based on the DCoDR paper\cite{kahana2022contrastive} with \textit{SOFTWARE RESEARCH LICENSE
} detailed here\footnote{https://github.com/jonkahana/DCoDR/blob/main/LICENSE}. The implementation uses the \textit{PyTorch} and \textit{faiss} \cite{johnson2019billion} packages.  PyTorch Uses a BSD-style license, as detailed in their license file\footnote{https://github.com/pytorch/pytorch/blob/master/LICENSE}. \textit{faiss} uses \textit{MIT License}.

The CLIP\cite{radford2021learning} network we used for automatic labelling uses \textit{MIT License}.

SimCLR \cite{chen2020simple} used by DcoDR and as a baseline uses \textit{Apache License}.

\end{document}